\documentclass[manuscript,screen]{acmart}

\usepackage{amsmath,amsfonts}
\usepackage{algorithmic}
\usepackage{algorithm}

\AtBeginDocument{%
  \providecommand\BibTeX{{%
    \normalfont B\kern-0.5em{\scshape i\kern-0.25em b}\kern-0.8em\TeX}}}

\begin{document}

\title{Two-stream Multi-level Dynamic Point Transformer for Two-person Interaction Recognition}

\author{Yao Liu}
\authornote{Both authors contributed equally to this research.}
\authornote{Corresponding author.}
\email{yao.liu3@unsw.edu.au}
\author{Gangfeng Cui}
\authornotemark[1]
\email{g.cui@student.unsw.edu.au}
\affiliation{
  \institution{School of Computer Science and Engineering, University of New South Wales}
  \city{Sydney}
  \state{NSW}
  \postcode{2052}
  \country{Australia}
}

\author{Jiahui Luo}
\affiliation{
  \institution{School of Computing and Information Systems, University of Melbourne}
  \city{Melbourne}
  \state{Victoria}
  \postcode{3010}
  \country{Australia}
}
\email{jialuo7@student.unimelb.edu.au}

\author{Xiaojun Chang}
\affiliation{
  \institution{Faculty of Engineering and Information Technology, University of Technology Sydney}
  \city{Sydney}
  \state{NSW}
  \postcode{2007}
  \country{Australia}
}
\email{xiaojun.chang@uts.edu.au}

\author{Lina Yao}
\affiliation{
  \institution{Data 61, CSIRO and School of Computer Science and Engineering, University of New South Wales}
  \city{Sydney}
  \state{NSW}
  \postcode{2015}
  \country{Australia}
}
\email{lina.yao@unsw.edu.au}

\renewcommand{\shortauthors}{Liu and Cui, et al.}

\begin{abstract}
As a fundamental aspect of human life, two-person interactions contain meaningful information about people's activities, relationships, and social settings. 
Human action recognition serves as the foundation for many smart applications, with a strong focus on personal privacy. However, recognizing two-person interactions poses more challenges due to increased body occlusion and overlap compared to single-person actions. 
In this paper, we propose a point cloud-based network named Two-stream Multi-level Dynamic Point Transformer for two-person interaction recognition. 
Our model addresses the challenge of recognizing two-person interactions by incorporating local-region spatial information, appearance information, and motion information.
To achieve this, we introduce a designed frame selection method named Interval Frame Sampling (IFS), which efficiently samples frames from videos, capturing more discriminative information in a relatively short processing time.
Subsequently, a frame features learning module and a two-stream multi-level feature aggregation module extract global and partial features from the sampled frames, effectively representing the local-region spatial information, appearance information, and motion information related to the interactions.
Finally, we apply a transformer to perform self-attention on the learned features for the final classification. 
Extensive experiments are conducted on two large-scale datasets, the interaction subsets of NTU RGB+D 60 and NTU RGB+D 120. The results show that our network outperforms state-of-the-art approaches in most standard evaluation settings.

\end{abstract}

\ccsdesc[500]{Computing methodologies~Activity recognition and understanding}

\keywords{Two-person interaction recognition, Point cloud-based method, Frame sampling, Two-stream multi-level feature aggregation, Transformer}


\maketitle

\section{Introduction}
\begin{figure}[t]
\centering
\includegraphics[width=0.6\columnwidth]{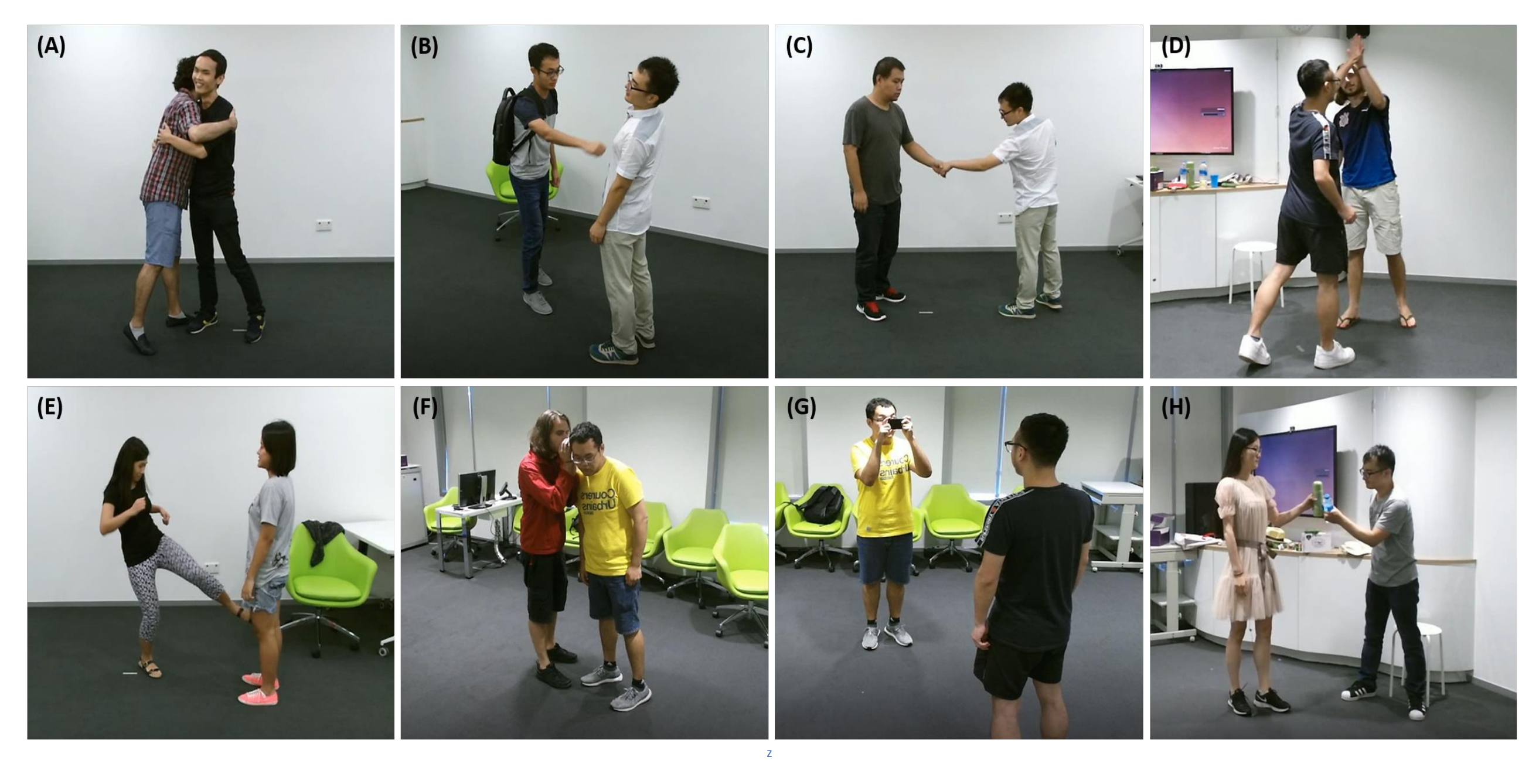} 
\caption{Examples of interactions from the NTU RGB+D 120 dataset: (A)~hugging, (B)~punching, (C)~shaking hands, (D)~high-five, (E)~kicking, (F)~whispering, (G)~taking a photo and (H)~cheers and drink.}
\label{fig_example}
\end{figure}

Interactions between two people, such as handshaking and hugging (see Figure~\ref{fig_example}), represent a fundamental aspect of human life. 
Recognizing such interactions is crucial and stands as one of the most significant branches of human activity recognition. Its applications span across various domains, including security, video retrieval, surveillance, and human-computer interfaces, promising substantial societal benefits\cite{Ahhi}.
A typical computer vision-based two-person interaction recognition task involves automatically identifying human interactions from image sequences or videos.
Despite notable advancements in human activity recognition over the past decade~\cite{acro,Continuous}, recognizing two-person interactions remains challenging.
Unlike single-person action recognition, identifying two-person interactions is more complex, primarily due to the involvement of multiple human subjects and the interdependence between them.

Earlier human action recognition heavily relied on wearable devices, which captured motion information to achieve action recognition \cite{Wearables}. 
However, this method required individuals to wear sensors continuously, making it costly for large-scale data collection and hindering its widespread adoption. 
In recent years, RGB-D-based two-person interaction recognition \cite{rgbain} has garnered significant attention, mainly due to the advancement of cost-effective RGB-D sensors. Several approaches in this domain focus on utilizing skeleton data alone \cite{tomm-skeleton1, tomm-skeleton2, tomm-skeleton3} or hybrid features from different channels \cite{tomm-multi}.
While these techniques can provide valuable information for recognition, they face critical challenges. 
For skeleton-based approaches, they are common in single-person action recognition, but there are still challenges in human-to-human interaction recognition \cite{rgbdhm}.
There are more body overlaps and occlusions in human-human interactions.
During the conversion from the RGB-D data to obtain the skeleton data, errors in the skeleton estimation can accumulate errors in action recognition and lead to performance degradation \cite{3dv}.
In addition, the skeleton-based approach requires pre-designed skeleton points, which requires extra work for two-person interaction recognition.
Regarding hybrid feature-based methods, the combination of features from multiple channels can result in heavy computations. Typically, these channels include RGB data\cite{tomm-local, tomm-video}, which contains texture and color information, raising concerns about privacy issues, particularly in scenarios such as monitoring the activities of children and older individuals in a room. Action analysis through common video or image data becomes challenging in such privacy-sensitive cases.

In this work, we propose a novel network named the Two-stream Multi-level Dynamic Point Transformer (TMDPT) for two-person interaction recognition, utilizing depth videos as the sole input to the network. 
Depth videos can be captured using depth cameras and are also accessible through open datasets collected by RGB-D cameras. Unlike RGB data, depth videos lack texture and color features and are insensitive to illumination, ensuring a higher level of personal privacy. 
Moreover, our model directly performs action recognition from the raw data, making it more robust against occlusion issues encountered by skeleton-based methods. 
Taking inspiration from the recent success of point cloud-based action recognition approaches \cite{3dv,seq}, we convert depth videos into point cloud videos in this paper. An additional advantage of this approach is the reduced computation costs when analyzing point cloud videos compared to processing depth videos directly.

In interaction learning, frame sampling plays a critical role due to the high frame rate of videos, typically containing a dozen frames per second.
Some adjacent or nearby frames have a large amount of repetitive content, which belongs to the redundancy information.
A well-designed frame sampling method is essential to balance efficiency and accuracy. 
However, in current interaction recognition methods, frame sampling methods are often not thoroughly investigated, and many simply employ uniform or random downsampling \cite{seq}. 
Such simplistic approaches may lead to performance degradation as selecting a small frame rate could cause the networks to miss crucial partial information, while opting for a large value could introduce unnecessary computational burden. In this paper, we introduce a novel frame sampling technique called Interval Frame Sampling (IFS), which intelligently selects informative frames while removing redundancy from the original videos, ultimately enhancing recognition performance.

With the sampled frames, we construct an effective deep-learning model for interaction recognition. 
Firstly, following the approach in \cite{seq}, we generate frame-level features. 
Subsequently, we propose a novel Two-stream Multi-level Feature Aggregation module to jointly capture informative global and partial features. This module comprises a global stream and a partial stream. 
Within the partial stream, we implement a temporal split procedure to divide a video into multiple temporal segments, allowing for extraction of partial information. In both global and partial streams, we carefully design and extract local-region spatial information, appearance information, and motion information, effectively representing the crucial aspects of two-person interactions and addressing the challenges associated with interaction recognition.
To further enhance the feature representations, we employ a transformer \cite{atte}, leveraging self-attention on the learned global and partial features. This process enables the global feature to acquire finer partial motion and appearance details while empowering the partial features with a global perspective. The output of the transformer, in combination with the original global feature, is then used for the final classification.

The main contributions of our work include the following:
\begin{itemize}
\item 
We propose a novel network named the Two-stream Multi-level Dynamic Point Transformer (TMDPT) for two-person interaction recognition. Notably, we are the pioneers in addressing interaction recognition tasks through depth video using a point cloud-based approach. This novel perspective expands the possibilities of depth-based interaction analysis.
\item 
We introduce the Interval Frame Sampling (IFS) technique for frame sampling, and leverage it to extract global and partial features through our two-stream multi-level feature aggregation module. By doing so, we effectively capture and represent local-region spatial information, appearance information, and motion information, all of which contribute to a comprehensive representation of two-person interactions.
\item 
Our network outperforms state-of-the-art approaches in most of standard evaluation settings of the interaction subsets of the NTU RGB+D 60 and NTU RGB+D 120 datasets.
This remarkable performance improvement validates the effectiveness of our proposed TMDPT model and its potential for practical applications.
In addition to achieving state-of-the-art results, we conduct an extensive ablation study to demonstrate the effectiveness of each module in our network. This analysis further reinforces the contributions and benefits of our proposed method.
\end{itemize}

\section{Related Works}

\subsection{Two-person interaction recognition}
Most recent two-person interaction recognition approaches heavily rely on RGB-D data \cite{rgbain}. 
These methods can be broadly classified into two major groups: skeleton-based interaction recognition and hybrid feature-based interaction recognition.

Skeleton-based methods are centered around utilizing skeleton data for interaction reasoning. This type of data contains 3D positions of body joints and is typically estimated from depth images. 
Various approaches within this category attempt to generate features based on each person's joints and their interactive joints to represent the motion relation.
For instance, in \cite{tpid}, an approach is proposed to use movements between all pairs of joints, joint distances, and velocity information as the motion information for interaction recognition.
Another study, \cite{RDT}, extracts a human interaction feature descriptor encompassing the directional, static, and dynamic properties of the skeleton data. They subsequently employ a linear model with a sparse group lasso penalty enhancement to facilitate the interaction recognition task.
Meanwhile, some researchers transform the interaction problem into a single-person action recognition problem. In \cite{htl}, a method is proposed to decompose a two-person interaction into two individuals' actions, followed by separate classification of each person's action.
However, a significant challenge in using skeleton data is the robustness of the skeleton estimation, which is still not a trivial matter. Inaccurate and incomplete estimated data can adversely affect recognition performance, thereby posing a serious limitation to the effectiveness of skeleton-based methods.

Hybrid feature-based methods leverage combined features from different channels to recognize interactions. 
For instance, \cite{rcar} integrates motion features, skeleton joints-based postures, and local appearance features from both depth and RGB data to reason about interactions.
Similarly, \cite{rmmc} combines the distance property of the 3D skeleton with dense optical features extracted from RGB and depth images jointly to predict the interaction class. 
On the other hand, \cite{DID} merges information from joints with poselets to select important frames and captures motion and appearance features for each interactive person from the bounding boxes where the human subjects are located.
Although features from multimodal sources can provide valuable information for recognition, these methods often require substantial amounts of data, affecting the network's efficiency. Additionally, many of these methods involve using RGB images as part of the input, which could expose excessive personal privacy information. Consequently, these methods may not be suitable for practical applications where some users prefer unobtrusive monitoring.

\subsection{Point cloud-based deep learning methods}
Point clouds contain abundant 3D geometrical information. 
PointNet \cite{pointnet} is a groundbreaking work that employs deep learning techniques to address static point cloud problems, including object part segmentation, classification, and semantic scene segmentation. The primary concept behind this approach is to utilize a symmetric function constructed with shared-weight deep neural networks to extract information from the points. Building on this, PointNet++ \cite{pointnet+} is a subsequent work that captures spatial features from local partitions and then hierarchically merges them to form a frame-level global feature.
The subsequent 3DGCN \cite{3DGCN} and IGCN \cite{IGCN} improve the performance of the model by defining variable kernels and interpolating kernels for point cloud feature extraction, respectively.

Compared to static point cloud analysis, learning on dynamic point clouds presents more challenges. 
Recently, various point cloud-based approaches have been devised to address video-level human action recognition tasks. 
For instance, in the 3DV method \cite{3dv}, 3D points are initially converted into regular voxel sets to capture motion information. These voxel sets are then abstracted back to a point set, serving as the input to PointNet++ \cite{pointnet+} for action recognition. Similarly, SequentialPointNet \cite{seq} flattens a point cloud video into a new data type named hyperpoint sequence, followed by feature mixing on the hyperpoint sequence to extract appearance and motion information.
The advantage of using point clouds is that they can be directly derived from depth images, thereby mitigating the exposure of excessive texture and color information found in RGB images and effectively safeguarding privacy. Moreover, depth images aid in distinguishing between background and people, and employing advanced point cloud processing techniques can reduce computational costs and enhance efficiency.
Taking inspiration from these methods, our paper proposes an effective point cloud-based network for two-person interaction recognition.

\subsection{Impact of frame rate}

In video data processing, it is common to convert the video into a sequence of frame-by-frame images for further analysis.
Despite the extensive research conducted in the area of human activity recognition \cite{acro}, the impact of frame rates has received relatively little attention until recently. 
\cite{fr_ra} is the first work to focus on this aspect. According to their findings, a lower frame rate (with fewer frames) results in a shorter running time, but it sacrifices the amount of information available for feature extraction, which can compromise the discriminability of the extracted features. On the other hand, a higher frame rate captures more temporal information, but it comes at the cost of longer processing times and potential redundancy. This excess redundancy can negatively affect the networks by introducing distracting effects.
As a result, employing a simple frame sampling method to evenly or consecutively select frames with either a lower or higher frame rate may not be optimal. The frame rate and the sampling method of the video have a significant impact on subsequent tasks, and it is essential to carefully consider and choose the appropriate frame rate and sampling technique for optimal performance.

\subsection{Transformer}
The Transformer architecture \cite{atte} was initially developed to address language-related problems, such as text classification, machine translation, and question answering, by utilizing a self-attention mechanism to learn the relationships between elements in a sequence.

In recent years, the computer vision community has embraced transformer networks and adapted them for various vision-related tasks, achieving remarkable success. 
Various approaches have been created to tackle different computer vision problems, including image generation, object detection, segmentation, image recognition, video understanding, and visual question answering \cite{TiV}. For instance, \cite{Tr_ir} introduces the Vision Transformer (ViT) for image classification. The method splits an image into multiple fixed-size patches and utilizes their embeddings along with positional information as input for a transformer. 
The transformer's output is then used for classification.
MViT \cite{MViT} introduces the concept of multiscale in ViT, constructing the model by gradually expanding the channel capacity and reducing spatio-temporal resolution. This approach addresses the pyramid structure through a multi-headed attention mechanism for recognition tasks in both images and videos.
TimeSFormer \cite{TimeSFormer} presents a convolution-free model proposing divided attention based on ViT. By extending ViT to video processing, TimeSFormer treats videos as sequences of patches extracted from frames.
SequentialPointNet\cite{seq} is a model for 3D human action recognition, processing point cloud sequences through two modules: the intra-frame appearance encoding module for spatial structure and the inter-frame motion encoding module for temporal changes. Transformer is used in the temporal position embedding layer to capture temporal features.
P4Transformer~\cite{P4T} employs a point 4D convolution to understand spatio-temporal structures and a transformer mechanism for grasping comprehensive appearance and motion details across videos, using self-attention on localized features instead of explicit tracking.
Building on the success of these methods, we integrate a transformer into our network to enhance the information contained in global and partial features, thereby improving performance in two-person interaction recognition.

\begin{figure*}[t]
\centering
\includegraphics[width=0.9\linewidth]{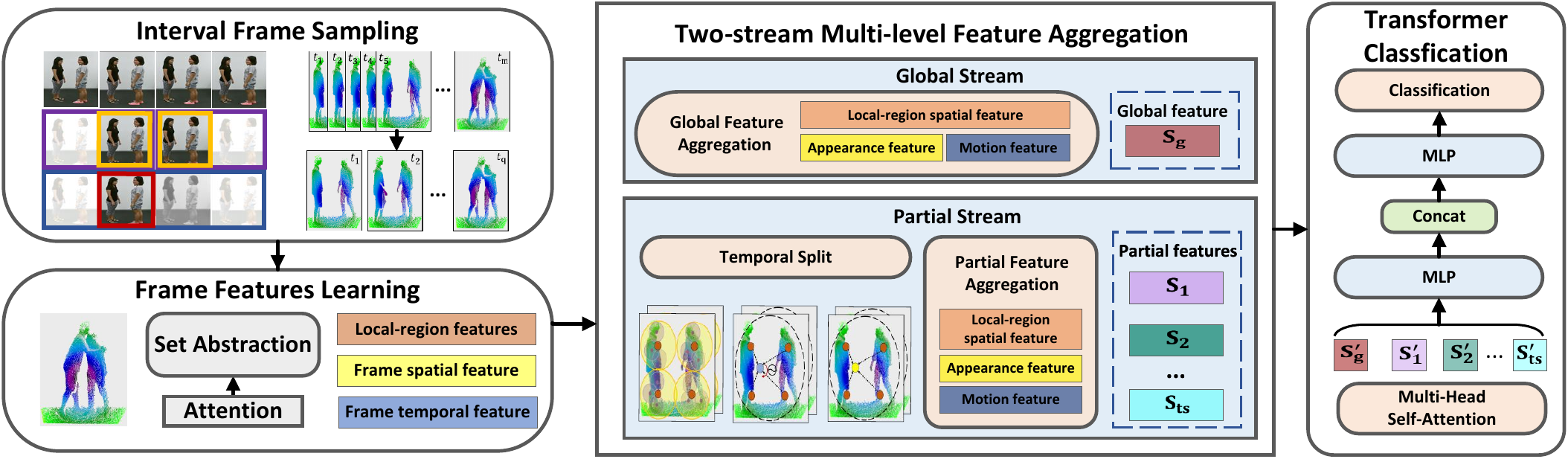} 
\caption{Two-stream Multi-level Dynamic Point Transformer comprises four main components: a novel frame sampling scheme named Interval Frame Sampling, a frame features learning module, a two-stream multi-level feature aggregation module, and a transformer classification module.}
\label{fig_overview}
\end{figure*}

\section{Methodology}

The overview of our Two-stream Multi-level Dynamic Point Transformer (TMDPT) is depicted in Figure~\ref{fig_overview}. TMDPT comprises four main components: a novel frame sampling scheme, a frame features learning module, a feature aggregation module, and a transformer classification module. 
We solely utilize the depth videos channel captured by the RGB-D camera as the input to our model. While the figure occasionally includes RGB picture depictions for illustrative purposes, please note that our model does not involve any RGB channels. 
First, we form 3D voxels by mapping depth values to coordinates, and convert the 3D grid into a point cloud by determining whether the voxels are occupied \cite{3dv,seq}.
Subsequently, Interval Frame Sampling (IFS) is applied to sample frames from the point cloud videos. The frame features learning module then extracts local-region features, frame spatial features, and frame temporal features from each frame. 
Next, a two-stream multi-level feature aggregation module combines these features into global and partial features. 
Lastly, the transformer classification module employs self-attention on the aggregated features and utilizes the output in conjunction with the original global feature for the final classification.

\subsection{Interval Frame Sampling}
\begin{figure*}[t]
\centering
\includegraphics[width=1\linewidth]{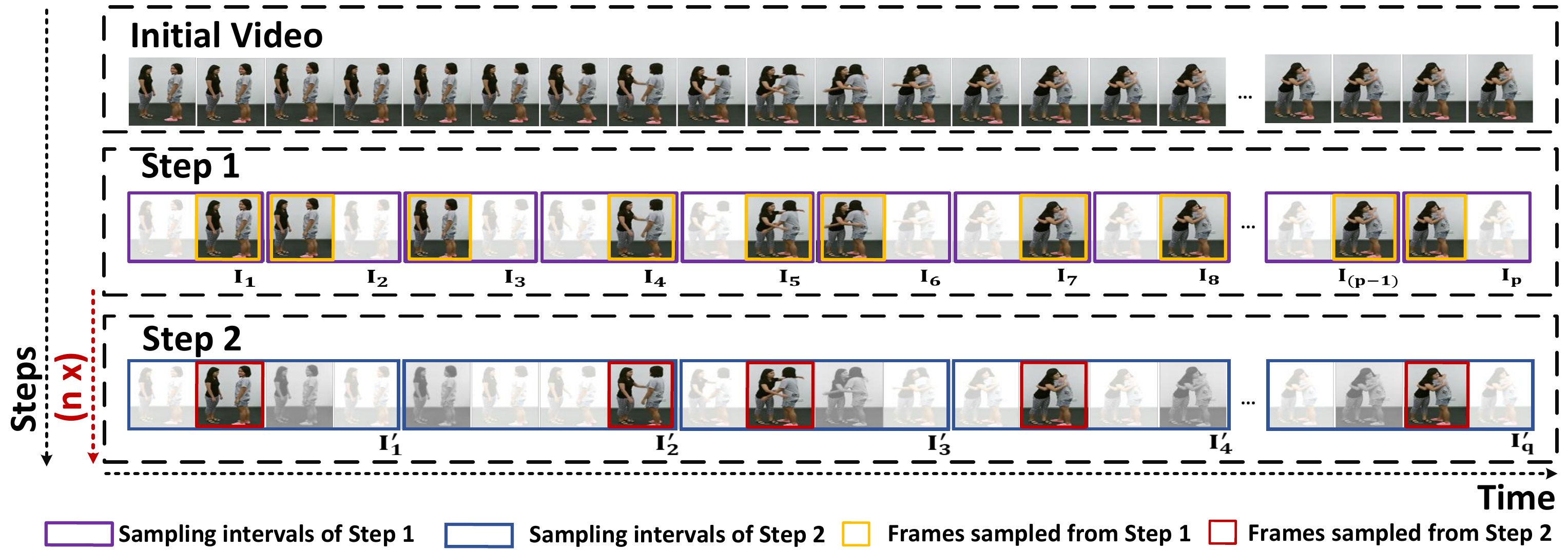} 
\caption{A schematic overview of Interval Frame Sampling for two-person interaction recognition. Starting with an original interaction video, in step 1, $p$ frames are sampled from $p$ intervals to represent the corresponding interaction sample during the whole learning process. In step 2, $q$ frames are further sampled from the $p$ frames for each epoch's interaction learning.}
\label{fig_ifs}
\end{figure*}

We introduce a novel frame sampling technique called Interval Frame Sampling (IFS) (refer to Figure~\ref{fig_ifs}). 
IFS comprises two steps, with the first step executed before the training phase to sample $p$ frames per interaction video from the original videos. 
To elaborate, given a point cloud video, we evenly separate all frames into $p$ intervals and then randomly select one frame from each interval. 
This step balances the advantages of uniform and random sampling to reduce redundancy while retaining highly discriminative information.
We denote $p$ as the Top frame rate in our method, usually selecting a large value to ensure that the interaction samples contain ample information. 
During the training phase, the second step is performed in each epoch. In this step, we apply the same sampling method again to evenly divide the updated video from the first step into $q$ intervals. We then randomly select one frame from each interval, resulting in a total of $q$ frames as a training instance. Essentially, this means the frames representing the same interaction sample are likely to vary from epoch to epoch. We refer to $q$ as the Bottom frame rate in this work.

Our frame selection technique offers several advantages. 
Firstly, by setting the interval size in the second step appropriately (not too large compared to the total training epochs), our network can effectively access a significant portion of the information contained within the $p$ frames. As mentioned earlier, $p$ can be set as a large number to provide our network with ample information.
Moreover, the two rounds of interval sampling within IFS help eliminate redundant information that may be present in consecutive interaction frames. This ensures that the network receives only essential and non-redundant data.
Additionally, IFS contributes to shorter processing times, as our network only utilizes $q$ frames per interaction sample during training. Here, $q$ can be set as a much smaller value compared to $p$.
Lastly, the characteristic of IFS introducing slight changes to the same interaction sample in each epoch's training offers positive data augmentation-like effects. This variation enables the network to focus on learning discriminative partial information from different frames during the training phase.
Overall, these advantages make IFS a valuable addition to our network, enhancing its efficiency and performance in two-person interaction recognition.

\subsection{Frame features learning module}

\begin{figure}[t]
\centering
\includegraphics[width=0.9\columnwidth]{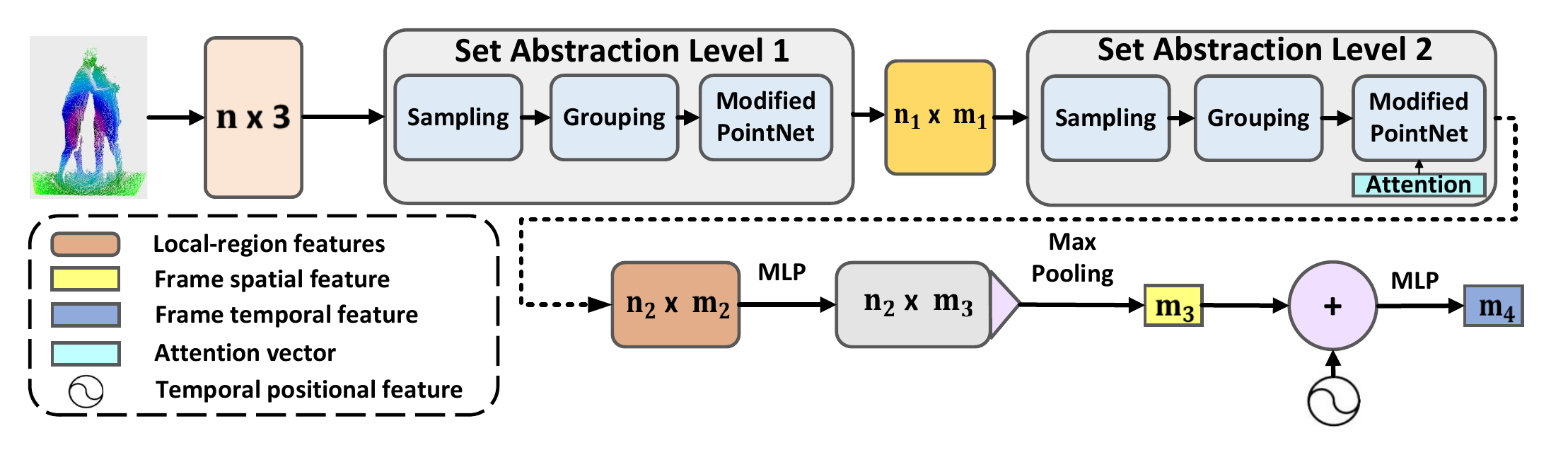}
\caption{A frame features learning module extracts local-region features, a frame spatial feature, and a frame temporal feature from each point cloud frame. $n$ denotes the number of points and $m$ denotes the dimension of the points.}
\label{fig_feat_learn}
\end{figure}

The frame features learning module outputs three types of features: local-region features, frame spatial features, and frame temporal features, which are obtained through a series of feature extraction steps (refer to Figure~\ref{fig_feat_learn}).
This module takes a point cloud set ${x_1, x_2, \dots, x_n}$ as input, where each $x_i \in \mathbb{R}^3$ represents the 3-dimensional coordinates of a point within the point cloud frame.

Specifically, the spatial feature extraction component consists of two levels of set abstraction (SA) \cite{pointnet+}. At each abstraction level, a sampling layer performs iterative Farthest Point Sampling (FPS) to select a fixed number of points as centroids from the input points. Subsequently, a grouping layer constructs local regions around these centroids using a ball query algorithm. Following that, a modified PointNet layer extracts local spatial features from each region.
In the modified PointNet layer, the coordinates of all points in each local region are transformed into local coordinates relative to the centroid point. Additionally, the distance between each point and its corresponding centroid is included as an additional point feature to enhance the network's performance in dealing with rotational motions. To further focus on learning crucial features, a Convolutional Block Attention Module (CBAM) \cite{CBAM} is applied to perform inter-feature attention on the point features. It is important to note that CBAM is not used in the first abstraction level within this work.

At the end of this layer, the feature vectors are combined with the coordinates of the corresponding centroid point, forming a spatial feature that represents a local region, which is referred to as local-region features.
Subsequently, a Multi-Layer Perceptron (MLP) and a Max Pooling operation are applied to capture a spatial feature representation for the entire frame, which we call the frame spatial feature.
The frame spatial feature extraction operation can be formulated as below:

\begin{equation}
\label{equ_frame_spa}
\begin{aligned}
FS_t = \underset{j=1,\dots,n_2}{MAX}\{MLP(e^t_j)\} = {MAX}\{MLP(SA(SA'(PC_t)))\}
\end{aligned}
\end{equation}

In Equation~\ref{equ_frame_spa}, $ FS_t $ is the frame spatial feature of the $ t\text{-th} $ point cloud frame ($ PC_t $). $ e^t_j $ is the abstracted feature of the $ j\text{-th} $ local region from the $ t\text{-th} $ frame. SA is the set abstraction operation.

After extracting the frame spatial features, a frame temporal feature extraction structure is constructed to capture temporal clues for each frame. In this step, a temporal positional feature vector, which contains the temporal positional information for each frame, is generated using a sinusoidal positional encoding technique \cite{atte}. Each dimension of the vector is calculated as follows:

\begin{equation}
\label{equ_pos_t1}
    TP(t,2j) = sin(\frac{t}{10000^{2j/m3}})
\end{equation}

\begin{equation}
\label{equ_pos_t2}
    TP(t,2j+1) = cos(\frac{t}{10000^{2j/m3}})
\end{equation}

In Equation~\ref{equ_pos_t1} and ~\ref{equ_pos_t2}, $ t $ is the temporal position. $ 2j $ and $ 2j+1 $ represent the dimension serial number. $ m3 $ is the size of a frame spatial feature. Each dimension of the sinusoidal positional vector corresponds to a sinusoid with a period of $10000^{2j/m3} \times 2\pi$. 
For a given input feature, this technique can be used to generate an absolute positional vector.

The generated temporal positional feature vectors have the same dimension as the frame spatial features; hence these two can be summed. After the summation, a MLP is applied to extract frame temporal features:

\begin{equation}
\label{equ_FT_t}
    FT_t = MLP(FS_t + TP_t)
\end{equation}

In Equation~\ref{equ_FT_t}, $ FT_t $, $ FS_t $ and $ TP_t $ are $ t\text{-th} $ frame temporal feature, frame spatial feature, and temporal positional feature, respectively.

\begin{figure}[t]
\centering
\includegraphics[width=0.8\columnwidth]{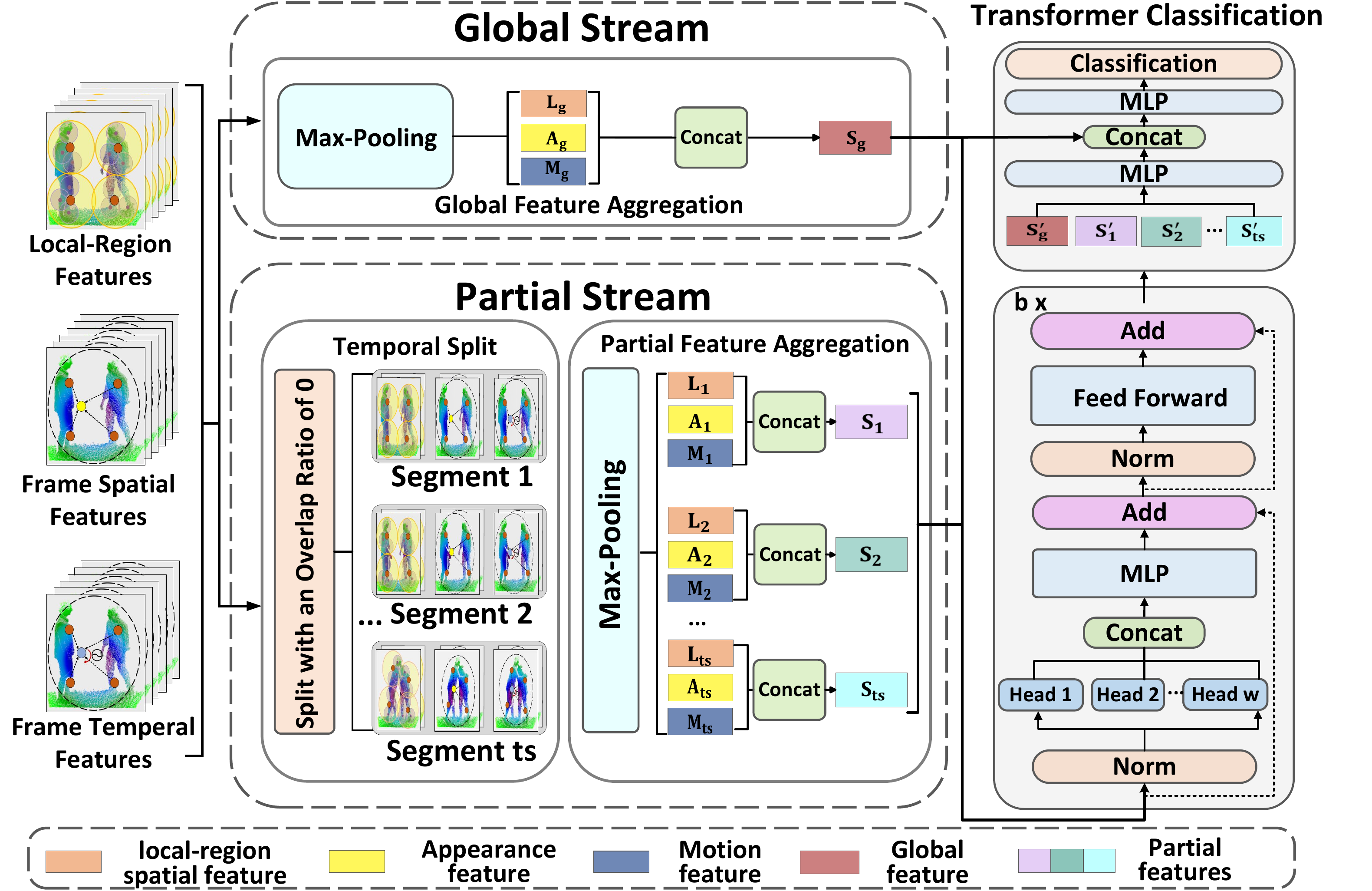}
\caption{First, a two-stream multi-level feature aggregation module merges frame-level features into global and partial features. Then, a transformer classification module performs self-attention on these aggregated features and uses the output combined with the original global feature for interaction recognition.}
\label{fig_two_multi}
\end{figure}

\subsection{Two-stream multi-level feature aggregation}\label{AQ}
With frame-level features generated, we propose a two-stream module to aggregate them into global and partial representations (the whole architecture with the following transformer classification module can be seen in Figure~\ref{fig_two_multi}).

\subsubsection{Two-stream feature aggregation} 

Both global and partial information is important for interaction recognition. 
Purely focusing on global information may cause a network to overlook crucial partial details necessary for distinguishing similar interactions. To address this, we propose a two-stream feature aggregation module, comprising a global stream and a partial stream, to jointly capture global and partial information.
Specifically, in the partial stream, a temporal split procedure is applied to obtain fine partial clues. Each video is split into $ts$ consecutive temporal segments, with an overlap ratio of 0.

\subsubsection{Multi-level feature aggregation} 

The multi-level feature aggregation technique has demonstrated its ability to enhance performance in tasks such as image segmentation and classification. In this work, we apply this technique to simultaneously aggregate local-region features, frame spatial features, and frame temporal features, providing our network with more discriminative information. 

Under the two-stream structure, our model processes both the temporal segments and the entire video. Initially, we utilize the multi-level feature aggregation technique to merge local-region features, frame spatial features, and frame temporal features, resulting in a local-region spatial feature, an appearance feature, and a motion feature. These features are then concatenated into a single feature, effectively describing the appearance and motion patterns jointly.
In total, our model generates one global feature and $ts$ partial features (corresponding to $ts$ temporal segments). Finally, all of these global and partial features are combined to form an integrated feature, enabling our network to leverage both global and fine-grained temporal information for accurate two-person interaction recognition.

\begin{equation}
\label{equ_S_g}
\begin{aligned}
S_g = Concat(L_g, A_g, M_g) = Concat(\underset{t = 1,\dots,q}{MAX} \{\underset{j=1,\dots,n_2}{MAX}\{ e^t_j \} \}, \underset{t = 1,\dots,q}{MAX} \{ FS_t \}, \underset{t = 1,\dots,q}{MAX} \{ FT_t \})
\end{aligned}
\end{equation}

\begin{equation}
\label{equ_S_i}
\begin{aligned}
S_i = Concat(L_i, A_i, M_i) = Concat(\underset{t = i,\dots,i+u-1}{MAX}\{\underset{j=1,\dots,n_2}{MAX}\{ e^t_j \} \} , \underset{t = i,\dots,i+u-1}{MAX}\{ FS_t \} , \underset{t = i,\dots,i+u-1}{MAX}\{ FS_t \}) 
\end{aligned}
\end{equation}

\begin{equation}
\label{equ_S}
    S = Concat(S_g, S_1,\dots,S_{ts})
\end{equation}

In Equation~\ref{equ_S_g},~\ref{equ_S_i} and ~\ref{equ_S},  $ S_g $ is the feature representation for the whole video, $ S_i $ is the feature representation for the $ i\text{-th} $ temporal segment, and $ S $ is an integrated feature vector containing both global and partial information. 
$ S_g $ is composed of three sub-features: $ L_g $ as the global aggregated local-region spatial feature, $ A_g $ as the global appearance feature, and $ M_g $ as the global motion feature. 
Each partial feature also consists of three sub-partial aggregated features. $ L_i $, $ A_i $, and $ M_i $ above denote the partial aggregated local-region spatial feature, the partial appearance feature, and the partial motion feature of the $ i\text{-th} $ temporal segment, respectively. 
$ e^t_j $ is the abstracted feature of the $ j\text{-th} $ local region from the $ t\text{-th} $ frame. It is the output of the second set abstraction level. 
$ u $ is the size of each segment after temporal splitting.

\subsection{Transformer classification}
We employ a transformer to perform self-attention on the aggregated global and partial features. This step aims to enhance each feature's information by allowing them to learn important relationships from other related features. 
Consequently, the partial features can gain a global view, and the global feature can capture more fine-grained partial motion and appearance details. 

The transformer applies a multi-head structure to improve its performance. It consists of $ n $ heads and receives the integrated feature $ S $ as the input.
Specifically, within an arbitrary head $ h_i $, the transformer uses three learnable matrices, $ M_{ki} $, $ M_{vi} $ and $ M_{qi} $, to transform each sub-feature within $ S $ into three vectors, representing a key, a value, and a query. All the keys, values, and queries are then packed together into matrices $ K_i $, $ V_i $, and $ Q_i $, respectively:

\begin{equation}
    K_i = SM_{ki}, \ V_i=SM_{vi}, \ Q_i = SM_{qi}
\end{equation}

After that, an attention weights matrix $ A_i $ containing the relationships between each pair of sub-features within $S$ is computed by using an attention function. 
Next, an updated feature vector $ U_i $ is generated by computing the cross product of the attention matrix $ A_i $ and the values $ V_i $:

\begin{equation}
    A_i = Attention(Q_i,K_i) = Softmax(\frac{Q_iK_i^T}{d_k^{1/2}})
\end{equation}

\begin{equation}
    U_i = A_iV_i
\end{equation}

The same calculation is performed in every head parallelly. $ n $ different updated feature vectors are generated. Eventually, a final updated vector $ U $ is computed by concatenating all of these n vectors:

\begin{equation}
    U = Concat(U_1,~\cdots , U_n)
\end{equation}

The updated vector after the transformer has the same shape as the input feature $S$. It consists of one updated global feature and $ts$ updated partial features, representing the enriched information from the self-attention process.
In addition to the multi-head structure, the transformer incorporates several other components, including residual connections, LayerNorms, and linear layers. Moreover, to enhance performance, $b$ identical transformer blocks are stacked together, enabling the model to capture more complex and higher-order relationships within the data.
Following the self-attention process, a subsequent MLP layer compresses these transformed features into a single feature. Subsequently, another MLP layer utilizes this single feature, combined with the original global feature, as the input to predict the interaction class of the video.
This residual-like structure contributes to making the model more stable and improves its ability to effectively learn and represent complex interaction patterns.

\begin{algorithm}[ht] 
\caption{Two-stream Multi-level Dynamic Point Transformer.}
\label{algo}
\begin{algorithmic}[1]
 \renewcommand{\algorithmicrequire}{\textbf{Input:}}
 \renewcommand{\algorithmicensure}{\textbf{Output:}}
 \REQUIRE Depth video
 \ENSURE  Classification result
 \\ \textit{Initialisation} :
  \STATE { Point cloud frames $\leftarrow$ $Segment$(Depth Video) }
  \FOR {each point cloud frame}
  \STATE { $Downsampling$(Point cloud frame) }
  \ENDFOR
  \STATE { IFS Step I  }
 \\ \textit{LOOP Process}:
  \STATE { IFS Step II  }
  \FOR {each point cloud frame}
  \STATE { Local-region feature $\leftarrow$  $SA$($SA'$(Point cloud frame))}
  \STATE { Frame spatial feature $\leftarrow$  $MLP$($MAX$(Local-region feature))}
  \STATE { Frame temporal feature $\leftarrow$  $MLP$(Frame spatial feature, Temporal positional feature)}
  \ENDFOR
  \IF {Global steam}
  \STATE Local-region spatial feature ($L_g$) $\leftarrow$ $MAX$(Local-region features)
  \STATE Appearance feature ($A_g$) $\leftarrow$ $MAX$(Frame spatial features)
  \STATE Motion feature ($M_g$) $\leftarrow$ $MAX$(Frame temporal features)
  \STATE Global feature ($S_g$) $\leftarrow$ $Concat$($L_g$,$A_g$,$M_g$)
  \ENDIF
  \IF {Partial stream}
  \FOR {Temporal split (ts) $\in [1,6]$}
  \STATE Local-region spatial feature ($L_{ts}$) $\leftarrow$ $MAX_{ts}$(Local-region features)
  \STATE Appearance feature ($A_{ts}$) $\leftarrow$ $MAX_{ts}$(Frame spatial features)
  \STATE Motion feature ($M_{ts}$) $\leftarrow$ $MAX_{ts}$(Frame temporal features)
  \STATE Partial feature ($S_{ts}$) $\leftarrow$ $Concat$($L_{ts}$,$A_{ts}$,$M_{ts}$)
  \ENDFOR
  \ENDIF
  \STATE { Classification result $\leftarrow$ $Classifier$($Transformer$($S_{g}$, $S_{1}$ \dots $S_{6}$) , $S_g$) }
 \RETURN Classification result 
 \end{algorithmic} 
 
\end{algorithm}

\subsection{Algorithm}

The overall process of our Two-stream Multi-level Dynamic Point Transformer is depicted in Algorithm~\ref{algo}.
Steps 1 to 5 pertain to the pre-processing data stage, primarily involving the conversion of the depth video into a collection of point cloud frames, downsampling of the point cloud frames, and the execution of IFS Step I.
From Step 6 onwards, the algorithm is executed in loops during the training process and only once during the testing process. Step 6 corresponds to IFS Step II.
Steps 7 to 11 represent the frame features learning module as described in Section 3.2.
Steps 12 to 25 depict our two-stream multi-level feature aggregation module, responsible for obtaining both global and partial features, as discussed in Section 3.3.
Finally, the last step is the classifier that utilizes Transformer in Section 3.4, employed to obtain the final recognition results.

\section{Experiments}

\subsection{Datasets}

\subsubsection{NTU RGB+D 60} 

NTU RGB+D 60 \cite{ntu60} is a large-scale dataset for 3D human activity understanding. It is collected from 40 distinct human subjects and contains 4 million frames and 56,000 samples. These samples are captured by Microsoft Kinect v2 from 80 different viewpoints. For a same action, 3 cameras are used simultaneously to capture 3 different views. NTU RGB+D 60 has a two-person interaction subset with 11 different classes. There are 10,428 interaction video samples in this subset.
To set up a standard for evaluating the results on the dataset, the authors have defined two interaction recognition evaluation settings: cross-subject evaluation and cross-view evaluation. Under the cross-subject evaluation setting, 40 human actors are separated into two groups, each with 20 actors. These two groups are used for training and testing, respectively. In the cross-view evaluation, the samples from camera 2 and camera 3 are used for training. The remaining samples from camera 1 are used for testing.

\subsubsection{NTU RGB+D 120} 

NTU RGB + D 120 \cite{ntu120} is the largest dataset for 3D action recognition. It extends NTU RGB+D 60 with a large amount of extra data. Compared with NTU RGB+D 60, it contains 60 additional classes. In total, it consists of more than 114,000 video samples and 8 million frames, which are collected from 106 different human subjects. 32 collection setups are used to build the dataset. Over different setups, the location and background are changed. It contains a two-person interaction subset with 24,828 video samples and 26 different classes. 
The authors have also suggested two standard evaluation settings for NTU RGB + D 120: cross-subject evaluation and cross-setup evaluation. The cross-subject evaluation shares the same rule as the one in NTU RGB+D 60, with a subset of human actors used for training and the rest used for testing. For the cross-setup evaluation, it has been defined to use the sample videos from even setup IDs for training and the ones from odd setup IDs for testing.

\subsection{Implementation details}
The Top and Bottom frame rates are set to 50 and 24, respectively, in the Interval Frame Sampling (IFS). 
That is saying 24 frames per video are used during training. 
We initially randomly sample 2048 points from each point cloud frame. Then, 512 points are further sampled from these 2048 points using a Farthest Point Sampling (FPS) method. 
Within the frame features learning module, FPS is applied again to select 128 centroids in the first set abstraction level and 32 centroids in the second level. 
In these two levels, the ball query radiuses are set to 0.06 and 0.1.
In the partial stream of the two-stream feature aggregation module, a point cloud video (containing 24 frames) is split into 6 consecutive temporal segments with an overlap ratio of 0. 
There are 4 frames in each temporal segment from which a partial feature is generated. 
We stack 5 identical transformer blocks together within the following transformer, each with 18 heads. 
We train our network for 90 epochs in total. 
Adam \cite{adam} is selected as the optimizer with a batch size of 32. 
We set the learning rate to start with 0.001 and decay with a rate of 0.5 every 10 epochs. A same data augmentation method from \cite{3dv} is applied in this work.

\begin{table}[t]
\caption{Comparison of recognition accuracy (\%) on the two-person interaction subset of NTU RGB+D 60.}
\begin{center}
\resizebox{0.5\columnwidth}{!}{
    \begin{tabular}{l c c c}
    \toprule
    \textbf{Methods} & \textbf{Year}& \textbf{Cross-Subject}& \textbf{Cross-View} \\
    \midrule
    ST-LSTM \cite{STLSTM} & 2016 & 83.0 & 87.3 \\
    GCA-LSTM \cite{GCA} & 2017 & 85.9 & 89.0 \\
    SPDNet \cite{SPDNet} & 2017 & 74.9 & 76.1 \\
    2S-GCA-LSTM \cite{2SGCA} & 2018 & 87.2 & 89.9 \\
    ST-GCN \cite{ST-GCN} & 2018 & 83.3 & 87.1 \\
    AS-GCN \cite{ASGCN} & 2019 & 89.3 & 93.0 \\
    2S-AGCN \cite{2SAGCN} & 2019 & 92.4 & 95.8 \\
    FSNET \cite{FSNET} & 2020 & 74.0 & 80.5 \\
    MS-AGCN \cite{MSAGCN} & 2020 & 94.1 & 96.7 \\
    3DV-PointNet++ \cite{3dv} & 2020 & 91.0 & 92.4 \\
    LSTM-IRN \cite{LSTMIRN} & 2021 & 90.5 & 93.5 \\
    SB-LSTM \cite{SBLSTM} & 2021 & 93.9 & 95.6 \\
    GeomNet \cite{GeomNet} & 2021 & 93.6 & 96.3 \\
    SequentialPointNet \cite{seq} & 2021 & 96.0 & 98.1 \\
    2s-AIGCN \cite{2022Attn} & 2022 & 95.3 & 98.0 \\
    3s-EGCN-IIG \cite{2022IIG} & 2022 & \textbf{96.6} & 98.7 \\
    JointContrast \cite{2023Joint} & 2023 & 94.1 & 96.8 \\
    \midrule
    TMDPT (Ours) &     2023 & \textbf{96.6} & \textbf{99.0} \\
    \bottomrule
    \end{tabular}
}
\label{tab1}
\end{center}
\end{table}

\subsection{Comparison with the state-of-the-art methods}\label{AG}
We compare our network’s performance to the state-of-the-art methods on the two large-scale 3D two-person interaction datasets, the interaction subsets of NTU RGB+D 60 and NTU RGB+D 120. The comparison results can be seen in Table~\ref{tab1} and Table~\ref{tab2}. 
Impressively, our network achieves remarkable results, consistently outperforming most other methods in the four standard evaluation settings of these two challenging datasets.

\begin{itemize}
\item ST-LSTM \cite{STLSTM}: The authors propose Spatio-temporal LSTM with trust gates for 3D human action recognition.
\item GCA-LSTM \cite{GCA}: The authors propose Global Context-aware Attention LSTM Networks for 3D action recognition.
\item SPDNet \cite{SPDNet}: The authors construct a Riemannian network structure with Symmetric Positive Definite (SPD) matrix nonlinear learning in a deep model for visual classification tasks.
\item 2S-GCA-LSTM \cite{2SGCA}: The authors propose a Two-stream Global Context-aware Attention LSTM network for human action recognition.
\item ST-GCN \cite{ST-GCN}: The authors propose Spatio-temporal Graph Convolutional Networks for automatic learning of spatio-temporal patterns in data.
\item AS-GCN \cite{ASGCN}: The authors introduce Actional Links in Graph Convolutional Networks to capture potential dependencies directly from actions, thus completing skeleton-based action recognition.
\item 2S-AGCN \cite{2SAGCN}: The authors propose Two-stream Adaptive Graph Convolutional Networks for skeleton-based action recognition. The topology of the graph can be learned end-to-end by the BP algorithm.
\item FSNET \cite{FSNET}: The authors focus on online action prediction for streaming 3D skeleton sequences. The authors introduce expanded convolutional networks to model the dynamics by sliding windows on the time axis.
\item ST-GCN-PAM \cite{PAM}: The authors propose a method based on Spatial-temporal Graph Convolutional Networks, which uses the pairwise adjacency matrix to capture the relationship of person-person skeletons.
\item MS-AGCN \cite{MSAGCN}: The authors propose Multi-stream Attention-enhanced Adaptive Graph Convolutional Neural Network for skeleton action recognition. Graph topologies can be learned uniformly or individually based on data in an end-to-end manner, and this data-driven approach increases the flexibility of the model.
\item 3DV-PointNet++ \cite{3dv}: The key to 3D Dynamic Voxel is to compactly encode the motion information in the depth video into a regular voxel set via a temporal rank pool, which is then transformed into point cloud data for processing.
\item LSTM-IRN \cite{LSTMIRN}: The authors propose an Interaction Relational Network that uses minimal prior knowledge about the structure of the human body to drive the network to identify the related body parts and use LSTM for inference.
\item SB-LSTM \cite{SBLSTM}: The Stacked Bidirectional LSTM uses two vectors to encode joint dynamics and spatial interaction information.
\item GeomNet \cite{GeomNet}: This is a method of two-person interaction recognition by using 3D skeleton sequences. Its key idea is to use Gaussian distributions to capture statistics on $R^n$ and those on the space of Symmetric Positive Deﬁnite (SPD) matrices.
\item SequentialPointNet \cite{seq}: The authors propose a frame-level parallel point cloud sequence network; specifically, the authors ﬂatten the point cloud sequence into hyperpoint sequences and then use the proposed Hyperpoint-Mixer module to process the spatial and temporal features of the frames.
\item 2s-AIGCN \cite{2022Attn}: The authors propose the Attention Interactive Graph Convolutional Network (AIGCN), which employs an Interactive Attention Encoding GCN (IAE-GCN) to extract interactive spatial structures and an Interactive-Attention Mask TCN (IAM-TCN) to discern temporal interactive features.
\item 3s-EGCN-IIG \cite{2022IIG}: The authors propose a method for recognizing interactions between two individuals by employing factorized convolution and analyzing the distances between joints.
\item JointContrast \cite{2023Joint}:The authors propose the Interaction Information Embedding Skeleton Graph Representation (IE-Graph) model, which utilizes unsupervised pre-training.
\end{itemize}

\subsubsection{NTU RGB+D 60} 

As the Table~\ref{tab1} shows, our network achieves impressive results of 96.6\% and 99.0\% on this large-scale dataset under the cross-subject and cross-view evaluation settings. Both results are the best among all these methods on this dataset. This shows that our network is robust against subject and view variation for interaction recognition tasks. One main reason for this success is that our network design enables our TMDPT to obtain crucial global and partial information to complete the interaction recognition tasks effectively.

\begin{table}[t]
\caption{Comparison of recognition accuracy (\%) on the two-person interaction subset of NTU RGB+D 120.}
\begin{center}
\resizebox{0.5\columnwidth}{!}{
    \begin{tabular}{l c c c}
    \toprule
    \textbf{Methods} & \textbf{Year}& \textbf{Cross-Subject}& \textbf{Cross-Setup} \\
    \midrule
    ST-LSTM \cite{STLSTM} & 2016 & 63.0 & 66.6 \\
    GCA-LSTM \cite{GCA} & 2017 & 70.6 & 73.7 \\
    SPDNet \cite{SPDNet} & 2017 & 60.7 & 62.1 \\
    2S-GCA-LSTM \cite{2SGCA} & 2018 & 73.0 & 73.3 \\
    ST-GCN \cite{ST-GCN} & 2018 & 78.9 & 76.1 \\
    AS-GCN \cite{ASGCN} & 2019 & 82.9 & 83.7 \\
    2S-AGCN \cite{2SAGCN} & 2019 & 86.1 & 88.1 \\
    FSNET \cite{FSNET} & 2020 & 61.2 & 69.7 \\
    ST-GCN-PAM \cite{PAM} & 2020 & 83.3 & 88.4 \\
    MS-AGCN \cite{MSAGCN} & 2020 & 87.7 & 90.5 \\
    3DV-PointNet++ \cite{3dv} & 2020 & 84.4 & 84.5 \\
    LSTM-IRN \cite{LSTMIRN} & 2021 & 77.7 & 79.6 \\
    SB-LSTM \cite{SBLSTM} & 2021 & 83.9 & 83.4 \\
    GeomNet \cite{GeomNet} & 2021 & 86.5 & 87.6 \\
    SequentialPointNet \cite{seq} & 2021 & 91.8 & 92.2 \\
    2s-AIGCN \cite{2022Attn} & 2022 & 90.7 & 90.7 \\
    3s-EGCN-IIG \cite{2022IIG} & 2022 & 92.4 & \textbf{95.5} \\
    JointContrast \cite{2023Joint} & 2023 & 88.2 & 88.9 \\
    \midrule
    TMDPT (Ours) & 2023 & \textbf{92.7} & 93.3 \\
    \bottomrule
    \end{tabular}
}
\label{tab2}
\end{center}
\end{table}

\subsubsection{NTU RGB+D 120} 

\begin{figure*}[htbp]
\centerline{
\includegraphics[width=1\linewidth]{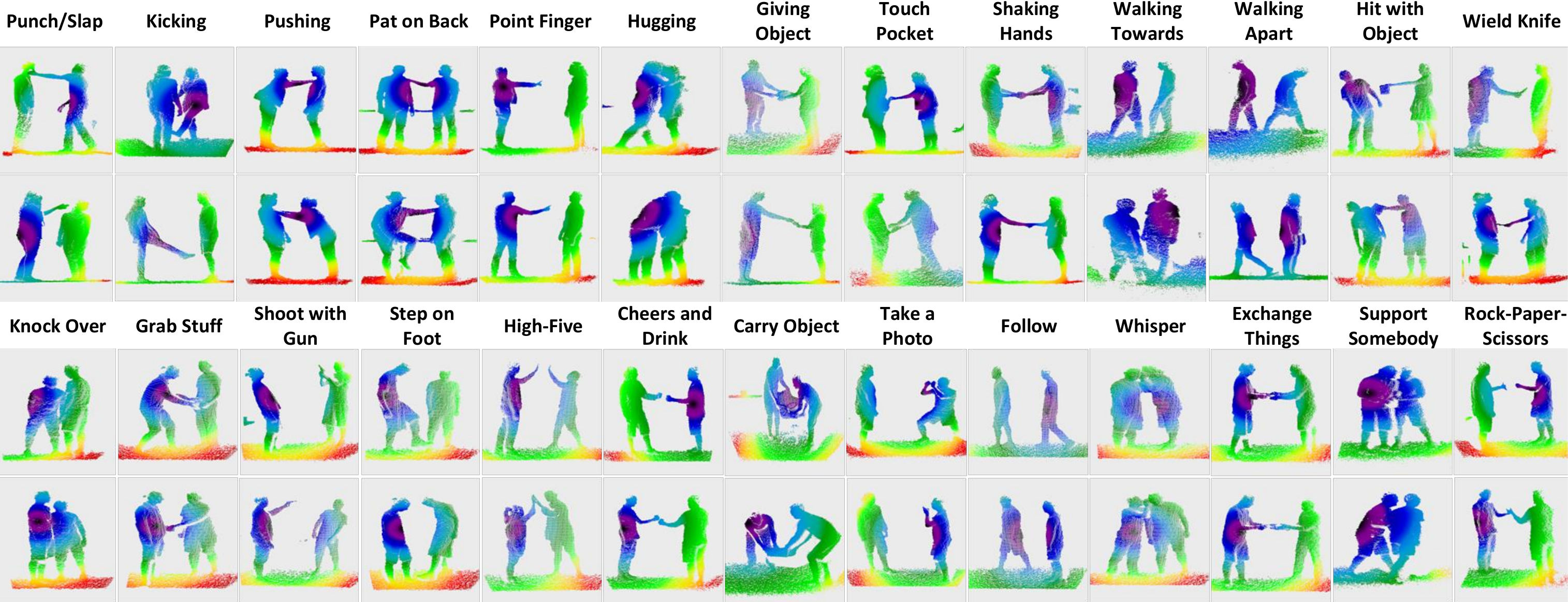}
}
\caption{Point cloud two-person interaction examples converted from the depth dataset of NTU RGB+D 120.}
\label{fig_vis}
\end{figure*}

Compared with NTU RGB+D 60, it is even harder to conduct interaction analysis on NTU RGB+D 120 (see point cloud interaction examples in Figure~\ref{fig_vis}). It is not only because NTU RGB+D 120 contains a larger amount of data than NTU RGB+D 60, but also because it has much higher variability in many different aspects, including action categories, subjects, camera views, and environments. As can be seen in Table~\ref{tab2}, despite facing more challenges, our network still achieves excellent results. With achieving an accuracy of 92.7\% on the cross-subject setting and 93.3\% on the cross-view setting, our network outperforms most of the state-of-the-art approaches again. This essentially demonstrates the effectiveness of our network for handling two-person interaction recognition tasks under complex subjects, environments, and viewpoints conditions.

\subsection{Ablation study}\label{AH}
A comprehensive ablation study is performed to extensively test the effectiveness of various aspects of our network design. All tests are conducted on NTU RGB+D 60 under the cross-view evaluation setting. 

\subsubsection{Effectiveness of Two-stream feature aggregation structure} 

In our network, a two-stream feature aggregation structure is proposed to aggregate global and partial information simultaneously. To evaluate its effectiveness, we create another network by removing the two-stream component with the following transformer and only using a global feature for the final classification. 
The performance comparison results can be seen in Table~\ref{tab3}. As the results show, the proposed two-stream structure can effectively improve the performance of our network by 1.17\%. Considering both results are close to 100\%, such a performance enhancement is indeed a notable improvement. This result demonstrates the effectiveness of our two-stream structure and reveals that both the global and partial information is important for interaction recognition. 

\begin{table}[htbp]
\caption{Effectiveness of two-stream feature aggregation structure on the
two-person interaction subset of NTU RGB+D 60 under cross-view test setting.}
\begin{center}
\resizebox{0.5\columnwidth}{!}{
    \begin{tabular}{l c}
    \toprule
    \textbf{Network Structure} & \textbf{Accuracy} \\
    \midrule
    Without Two-stream aggregation structure & 97.85 \\
    With Two-stream aggregation structure (proposed) & \textbf{99.02} \\
    \bottomrule
    \end{tabular}
}
\label{tab3}
\end{center}
\end{table}

\subsubsection{Effectiveness of Transformer} 

We propose to use a transformer to perform self-attention on the global and partial features. To demonstrate the effectiveness of the transformer, a test case with deleting it from our proposed network is created. We compare the performance of this new test case with our proposed one. From the results listed in Table~\ref{tab4}, we can observe that the recognition accuracy drops by 0.60\% without using the transformer. This shows that the self-attention process can bring meaningful extra information for interaction reasoning.

\begin{table}[htbp]
\caption{Effectiveness of transformer on the
two-person interaction subset of NTU RGB+D 60 under cross-view test setting.}
\begin{center}
\resizebox{0.35\columnwidth}{!}{
    \begin{tabular}{l c}
    \toprule
    \textbf{Network Structure} & \textbf{Accuracy} \\
    \midrule
    Without Transformer & 98.42 \\
    With Transformer (proposed) & \textbf{99.02} \\
    \bottomrule
    \end{tabular}
}
\label{tab4}
\end{center}
\end{table}

\subsubsection{Analysis of Transformer related design choices} 

To conduct transformer input analysis, we report the result of a new network, where the transformer only takes the global and partial motion features as input (without using the aggregated local-region spatial features and appearance features). The results of Table~\ref{tab5} demonstrate that performing self-attention on more comprehensive multi-level features (proposed) instead of only on the motion features can help our network achieve a better performance.
From the model structure, local-region features, frame spatial features and frame temporal features are output sequentially; however, for the aggregated features, local-region spatial features and appearance features still contain key features that are not in motion features, so the accuracy can be improved.

\begin{table}[htbp]
\caption{Comparison of recognition accuracy (\%) on the
two-person interaction subset of NTU RGB+D 60 under cross-view test setting with using different transformer inputs.}
\begin{center}
\resizebox{0.4\textwidth}{!}{
    \begin{tabular}{l c}
    \toprule
    \textbf{Transformer input} & \textbf{Accuracy} \\
    \midrule
    Motion features & 98.49 \\
    Multi-level features (proposed) & \textbf{99.02} \\
    \bottomrule
    \end{tabular}
}
\label{tab5}
\end{center}
\end{table}

Additionally, we investigate the performance variation of our network when using different feature aggregation methods (Max pooling and MLP) for merging the output of the transformer. As the results in Table~\ref{tab6} indicate, MLP (proposed) provides a better outcome than Max pooling. Simple Max pooling method is not involved in the learning process and cannot fit the data samples reasonably.
Thus, MLP is used as the transformer aggregation method in our network.

\begin{table}[htbp]
\caption{Comparison of recognition accuracy (\%) on the
two-person interaction subset of NTU RGB+D 60 under cross-view test setting with using different transformer output aggregation methods.}
\begin{center}
\resizebox{0.45\columnwidth}{!}{
    \begin{tabular}{l c}
    \toprule
    \textbf{Transformer Aggregation Methods} & \textbf{Accuracy} \\
    \midrule
    Max Pooling & 98.57 \\
    MLP (proposed) & \textbf{99.02} \\
    \bottomrule
    \end{tabular}
}
\label{tab6}
\end{center}
\end{table}

\subsubsection{Effectiveness of Interval Frame Sampling (IFS)}

In some advanced methods \cite{LSTMIRN, SBLSTM, GeomNet, seq}, they use a fixed frame rate to sample videos. 
To verify the effectiveness of Interval Frame Sampling (IFS), we test five different cases. In the first four cases, we use a traditional frame-selecting method to evenly sample fixed 6, 12, 24, and 50 frames per interaction video, respectively. Here, the frames representing a particular interaction video do not change over time during the training period. In the last case, we use our proposed IFS setting to choose 50 as the Top frame rate and 24 as the Bottom frame rate.

\begin{figure}[htbp]
\centerline{\includegraphics[width=0.9\linewidth]{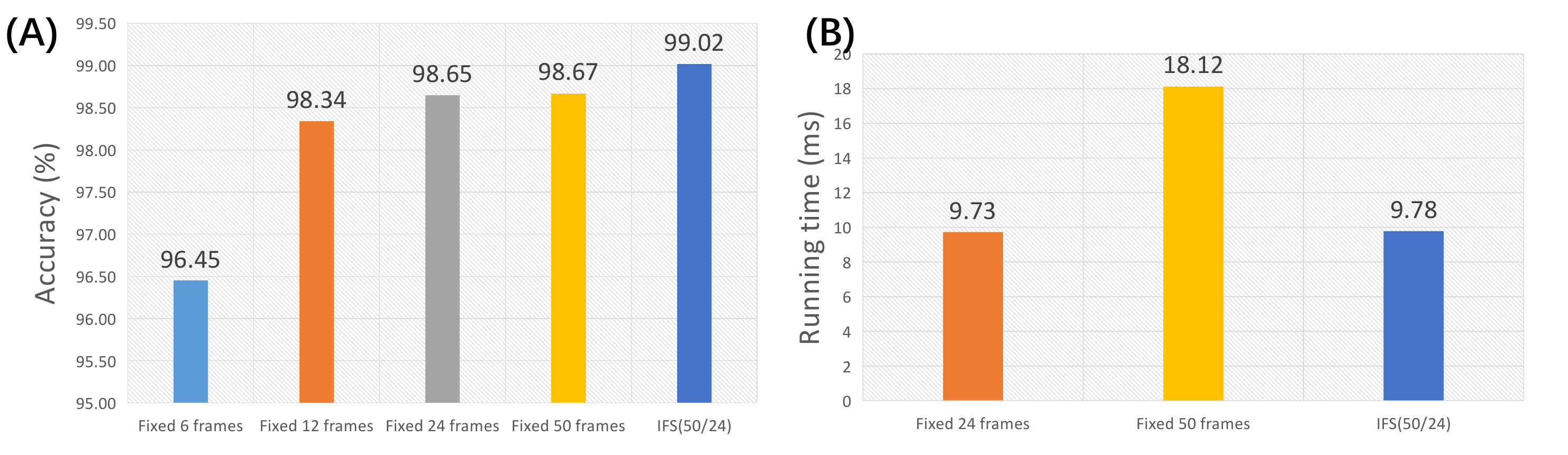}}
\caption{Comparison of frame sampling methods on the two-person interaction subset of NTU RGB+D 60 under cross-view test setting: (A) recognition accuracy (\%), (B) running time per video (ms).}
\label{fig_comp1}
\end{figure}

We first compare the recognition accuracy of these five cases. As shown in  Figure~\ref{fig_comp1} (A), more frames can lead to better performance when using the traditional frame selection methods. However, when the number of frames is larger than 24, the performance starts to be steady. This is because a higher frame rate can bring both extra helpful information and useless redundant information. While the former helps improve the performance, the latter can distract the network as noise. The result of our proposed approach indicates that this redundancy-related issue can be solved by our IFS. Compared with the two cases of using fixed 24 and 50 frames, we can gain an extra 0.37\% and 0.35\% accuracy increase with using IFS. These are considerable performance enhancements, considering the accuracy is already close to 100\%. 

We also compare the running time (per video) of using fixed 24 and 50 frames with our IFS,  Figure~\ref{fig_comp1} (B) shows that our IFS has a very similar processing time as the one using fixed 24 frames.

All these results demonstrate that IFS can improve the efficiency and accuracy of recognition effectively.
Our IFS benefits from both sides, having an accuracy of more than 50 frames, while the running time is equivalent to only 24 frames. This double advantage reflects the advancement of IFS.

\begin{figure}[htbp]
\centerline{\includegraphics[width=0.9\linewidth]{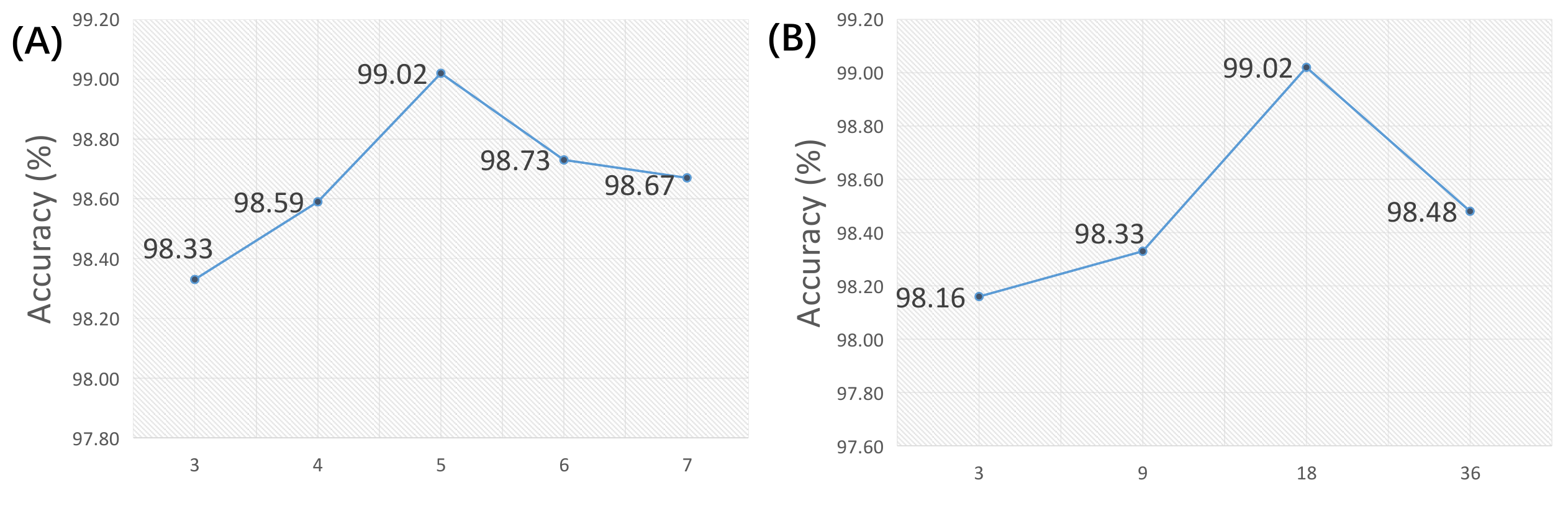}}
\caption{Recognition accuracy (\%) on the two-person interaction subset of NTU RGB+D 60 under cross-view test setting: (A)  different numbers of transformer blocks, (B)  different numbers of transformer heads per block.}
\label{fig_comp2}
\end{figure}

\subsubsection{Model hyperparameter analysis}

A series of model hyperparameter analysis is conducted in this work.
First, we investigate the impact of different numbers of blocks on the transformer. 
As shown in Figure~\ref{fig_comp2} (A), the accuracy is first improved when there are more blocks, reaching a peak when the transformer contains 5 blocks. 
Then, the accuracy starts to drop with additional blocks added. 
This result indicates that TMDPT can achieve a better performance with using multiple transformer blocks.
However, if there are too many blocks stacked together, issues like gradient exploding or vanishing can occur to bring adverse effects to the performance.

Then, we analyze the influence of the number of heads on the performance. Figure~\ref{fig_comp2} (B) shows that the multi-head design can improve TMDPT’s learning ability, as long as not too many heads are used. One main reason for this result is that a transformer with a reasonable number of heads can jointly attend to information from different representation subspaces at different positions to improve performance.
However, when there are too many heads, each head may not carry enough information to perform self-attention effectively.

\begin{table}[htbp]
\caption{Comparison of recognition accuracy (\%) on the
two-person interaction subset of NTU RGB+D 60 under cross-view test setting with using different temporal split methods.}
\begin{center}
\resizebox{0.6\columnwidth}{!}{
    \begin{tabular}{l c}
    \toprule
    \textbf{Temporal split methods} & \textbf{Accuracy} \\
    \midrule
    2 frames per segment with an overlap ratio of 0 & 98.21 \\
    6 frames per segment with an overlap ratio of 0 & 98.69 \\
    4 frames per segment with an overlap ratio of 0.5 & 98.36 \\
    4 frames per segment with an overlap ratio of 0 (Proposed) & {\bfseries 99.02} \\
    \bottomrule
    \end{tabular}
}
\label{tab11}
\end{center}
\end{table}

Finally, we test 4 cases with using different temporal split methods. As Table~\ref{tab11} shows, our proposed split method achieves the best result. Under this method, each video is split into 6 segments with an overlap ratio of 0. Too large or too small segmentation methods can affect the model's accuracy due to insufficient fitting or overfitting; also, we find that the segmentation does not need to overlap after our IFS sampling.
The result indicates that the proposed split method can help our network to extract more useful partial information for interaction learning.

\subsubsection{Model complexity analysis}

\begin{table}[htbp]
\caption{Model complexity analysis.}
\begin{center}
\resizebox{0.8\columnwidth}{!}{
    \begin{tabular}{l c c c c}
    \toprule
    \textbf{Methods} & \textbf{billion FLOPs} & \textbf{million parameters} & \textbf{rel. memory} & \textbf{rel. time}\\
    \midrule
    SequentialPointNet \cite{seq} & 54.80 & 2.98 &  2.61 & 2.75 \\
    TMDPT w/o Transformer (Ours) & 2.55 & 6.56 & 0.69 & 0.36 \\
    TMDPT (Ours) & 7.09 & 47.06 & 1 & 1 \\
    \bottomrule
    \end{tabular}
}
\label{tab_complex}
\end{center}
\end{table}

We perform a model complexity analysis of our TMDPT and compare it with SequentialPointNet \cite{seq}, which also uses a point cloud-based approach. 
As shown in Table~\ref{tab_complex}, our TMDPT model has lower FLOPs and higher number of parameters relative to SequentialPointNet.
Our model contains two-stream multilevel feature aggregation and the Transformer module thus significantly surpasses the SequentialPointNet model in terms of the number of parameters, but due to the sound design of our model network as well as the efficient operation and implementation, it results in a lead over the SequentialPointNet model in FLOPs.
In addition we also compare the resource consumption in real computation, SequentialPointNet is 2.61 times more expensive than our TMDPT in memory consumption and 2.75 times more expensive than our TMDPT in time consumption.
We find that the multi-head mechanism in the Transformer is responsible for affecting the overall number of model parameters. Therefore we have similarly compared the complexity of TMDPT w/o Transformer.
As a result, it can be found that a great saving of computational resources can be achieved without the use of a transformer. Comparing Tables~\ref{tab1} and ~\ref{tab4}, we find that on the two-person interaction subset of NTU RGB+D 60 under cross-view test setting, even without using a transformer, our TMDPT w/o Transformer~(98.4\%) exceeds the SequentialPointNet~(98.1\%).
This shows that in some scenarios with limited computational resources, leading results can still be obtained using the TMDPT w/o Transformer.

\subsubsection{Model robustness analysis}

\begin{table}[htbp]
\caption{Model robustness analysis on the two-person interaction subset of NTU RGB+D 60 under cross-view test setting.}
\begin{center}
\resizebox{0.6\columnwidth}{!}{
    \begin{tabular}{l c c}
    \toprule
    \textbf{Methods} & \textbf{512 points (accuracy \%)} & \textbf{256 points(accuracy \%)}\\
    \midrule
    SequentialPointNet \cite{seq} & 98.1 & 84.6\\
    TMDPT (Ours) & 99.0 & 96.8 \\
    \bottomrule
    \end{tabular}
}
\label{tab_rob}
\end{center}
\end{table}

Our model employs point clouds based on depth video transformations as input data, which avoids the errors accumulated in skeleton estimation compared to skeleton based methods.
However the number of points sampled from the point cloud data directly affects the performance of the model, here we compare the results of sampling 512 points (baseline) and 256 points.
Our experiments are performed on the two-person interaction subset of NTU RGB+D 60 under cross-view test setting and compare the results of the SequentialPointNet\cite{seq} model, which is also based on a point cloud approach. 
As seen in Table~\ref{tab_rob}, SequentialPointNet is only 0.9\% lower than our TMDPT when sampling 512 points, but SequentialPointNet is 12.2\% lower than our TMDPT when sampling 256 points.
This shows that our model maintains a certain performance with fewer point cloud samples and is more robust compared to SequentialPointNet.

\section{Conclusion}

In this study, we focus on the task of two-person interaction recognition and place a strong emphasis on privacy issues. To address this concern, we exclusively use depth video data for our research.
We propose a novel point cloud-based network named Two-stream Multi-level Dynamic Point Transformer (TMDPT) for two-person interaction recognition. TMDPT is composed of four main components: a unique frame sampling scheme, a frame features learning module, a two-stream multi-level feature aggregation module, and a transformer classification module.
Our novel frame sampling technique effectively selects key frames from the input video. The frame feature learning module then extracts frame-level features, while the two-stream multi-level feature aggregation module combines global and partial features.
Lastly, the classification module, which utilizes Transformer, performs two-person interaction classification. Our network is extensively evaluated on two large-scale 3D interaction datasets, specifically the interaction subsets of NTU RGB+D 60 and NTU RGB+D 120. The results of the experiments showcase the superiority of our proposed network for two-person interaction recognition.

\bibliographystyle{ACM-Reference-Format}
\bibliography{main}

\end{document}